\pdfoutput=1
\documentclass[11pt]{article}
\usepackage[preprint]{acl}
\usepackage{times}
\usepackage{latexsym}
\usepackage[T1]{fontenc}
\usepackage[utf8]{inputenc}
\usepackage{microtype}
\usepackage{inconsolata}
\usepackage{graphicx}

\title{{\sc \textbf{ImPart}}: Importance-Aware Delta-Sparsification \\for Improved Model Compression and Merging in LLMs}

\author{%
    Yan Yang$^{1*}$, \
    Yixia Li$^{2}$\thanks{\ \ Equal Contribution.}, \
    Hongru Wang$^{4}$, \
    Xuetao Wei$^{2}$ \ \\
    \textbf{Jianqiao Yu}$^{3}$, \ 
    \textbf{Yun Chen}$^{1}$, \
    \textbf{Guanhua Chen}$^{2}$ \\
    $^1$Shanghai University of Finance and Economics, $^2$Southern University of Science and Technology  \\
    $^3$Harbin Institute of Technology (Shenzhen), $^4$The Chinese University of Hong Kong \\
}

\usepackage{xspace}
\newcommand{\mname}{{\sc ImPart}\xspace}
\newcommand{\mnameqt}{{\sc ImPart-Qt}\xspace}
\newcommand{\mnameqtc}{IMPART-Qt}
\newcommand{\dareqt}{DARE-{\sc Qt}\xspace}

\usepackage{algorithm}
\usepackage[noend]{algpseudocode}
\usepackage{multirow}
\usepackage{makecell}
\usepackage{subcaption}
\usepackage{amsmath}
\usepackage{amsfonts}
\usepackage{amssymb}
\usepackage{booktabs}
\PassOptionsToPackage{table}{xcolor}
\usepackage{colortbl}
\usepackage{enumitem}

\begin{document}
\maketitle
\begin{abstract}
With the proliferation of task-specific large language models, delta compression has emerged as a method to mitigate the resource challenges of deploying numerous such models by effectively compressing the delta model parameters.
Previous delta-sparsification methods either remove parameters randomly or truncate singular vectors directly after singular value decomposition (SVD).
However, these methods either disregard parameter importance entirely or evaluate it with too coarse a granularity. In this work, we introduce {\sc ImPart}, a novel importance-aware delta sparsification approach. Leveraging SVD, it dynamically adjusts sparsity ratios of different singular vectors based on their importance, effectively retaining crucial task-specific knowledge even at high sparsity ratios. Experiments show that {\sc ImPart} achieves state-of-the-art delta sparsification performance, demonstrating $2\times$ higher compression ratio than baselines at the same performance level. When integrated with existing methods, {\sc ImPart} sets a new state-of-the-art on delta quantization and model merging.
\end{abstract}

\section{Introduction}
Large language models (LLMs) have demonstrated remarkable capabilities across diverse knowledge-intensive \cite{qwen2.5, abdin2024phi4technicalreport} and reasoning-intensive \cite{deepseekai2025deepseekr1incentivizingreasoningcapability, kimiteam2025kimik15scalingreinforcement} tasks through post-training. Different users fine-tune the widely applicable open-sourced base LLMs such as LLaMA \cite{grattafiori2024llama3herdmodels} and DeepSeek \cite{deepseekai2024deepseekv3technicalreport} with customized datasets for specific downstream tasks. However, maintaining separate fine-tuned models for each user or downstream task poses significant resource challenges \cite{ryu2023efficientstoragefinetunedmodels,yao2024deltazipefficientservingmultiple}, particularly in storage and deployment costs. The challenges have attracted increased interest within the community in efficient model compression techniques that can preserve task-specific knowledge while reducing resource requirements.

Recent approaches \cite{yu2024language, liu2024bitdelta, ping2024deltacome} propose to address this challenge by delta compression, which aims to compress the difference between the fine-tuned parameters and the base model parameters (i.e., delta parameters) by quantization or sparsification. 
Previous sparsification-based methods \cite{isik2023gptzip,yao2024deltazipefficientservingmultiple,yu2024language} sparsify the delta parameters by randomly setting partial weight entries to zero or truncating singular vectors directly after singular value decomposition (SVD). However, these methods fail to produce satisfactory results, particularly on challenging specialized tasks like math reasoning or code generation, as they inadvertently discard important parameters as the sparsification ratio increases.

\begin{figure*}[t]
    \vspace{-15pt}
    \centering
    \begin{subfigure}[b]{0.32\textwidth}
        \centering
        \includegraphics[width=\textwidth]{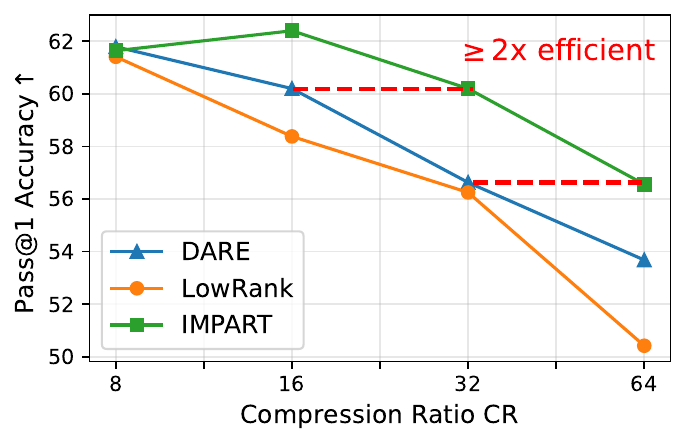}
        \caption{WizardMath-13B on GSM8K}
        \label{fig:diff_sparse_ratio_gsm8k}
    \end{subfigure}
    \hfill
    \begin{subfigure}[b]{0.32\textwidth}
        \centering
        \includegraphics[width=\textwidth]{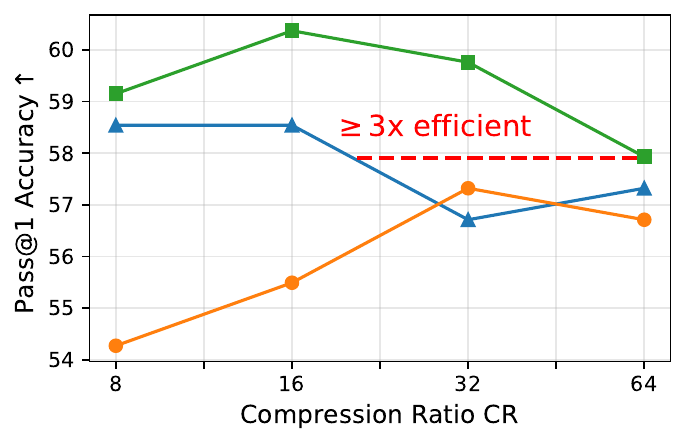}
        \caption{WizardCoder-13B on HumanEval}
        \label{fig:diff_sparse_ratio_humaneval}
    \end{subfigure}
    \hfill
    \begin{subfigure}[b]{0.32\textwidth}
        \centering
        \includegraphics[width=\textwidth]{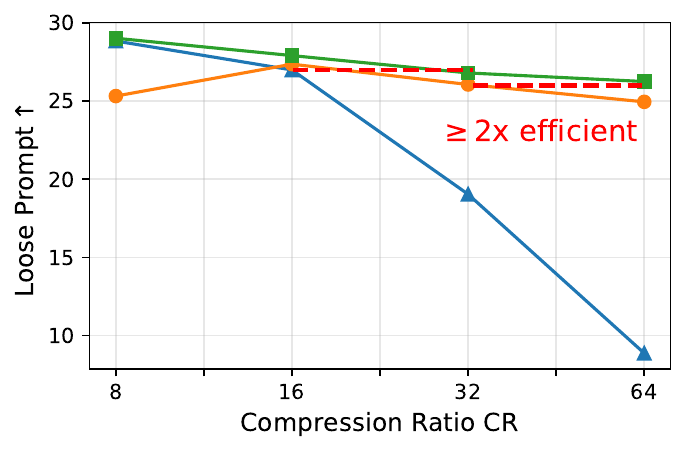}
        \caption{LLaMA2-Chat-13B on IFEval}
        \label{fig:diff_sparse_ratio_ifeval}
    \end{subfigure}
    \vspace{-8pt}
    \caption{Comparative evaluation of \mname against state-of-the-art sparsification methods across mathematical reasoning, code generation, and chat tasks. \mname consistently outperforms baselines across various tasks while maintaining high sparsity ratios (more detailed discussions are in Section~\ref{sec:diff_sparse_ratio}).}
    \label{fig:diff_sparse_ratio}
    \vspace{-15pt}
\end{figure*}

In this work, we propose \mname (\textbf{Imp}ortance-\textbf{A}ware Delta-Spa\textbf{r}sifica\textbf{t}ion), a novel effective sparsification-based delta compression approach even at high sparsity ratios. 
The \mname framework is motivated by the observations that singular vectors associated with larger singular values encode more important task-specific information \cite{ping2024deltacome,sharma2024the,ryu2023efficientstoragefinetunedmodels}.
Building on this insight, \mname proposes an adaptive sparsification mechanism that assigns different sparsity ratios on the singular vectors based on the corresponding singular values, ensuring the preservation of critical task-specific knowledge. Based on our theoretical analysis, the parameters of the sparsified singular vector are then re-scaled to ensure the performance is maintained.
The \mname framework can be applied to delta quantization or model merging tasks by integrating it with existing approaches thereby supporting a higher compression ratio. 

Extensive experiments on LLM sparsification across three diverse tasks with various backbones demonstrate the effectiveness of our method. As shown in Figure~\ref{fig:diff_sparse_ratio}, \mname demonstrates $2\times$ higher compression ratio than baselines at the same performance level, retaining 95.8\% of the fine-tuned model's performance at a compression ratio of 16 (93.75\% sparsity). Additional experiments on integration with quantization and model merging further validate \mname's versatility, making it a practical solution for deploying numerous fine-tuned language models in resource-constrained environments. \footnote{Our code are publicly available at \url{https://github.com/yanyang19/ImPart}.}

\section{Preliminaries} \label{sec:preliminary}

\paragraph{Delta Compression}
Delta parameters are the differences between the parameters of a fine-tuned LLM and its corresponding base LLM. In scenarios such as multi-tenant serving, where a large number of LLMs fine-tuned from the same base model are deployed to meet various and complicated user requirements, using $N$ sets of delta parameters in conjunction with the shared backbone can eliminate the need for $N$ full fine-tuned models.
Delta compression aims to compress these delta parameters by sparsification~\citep{yu2024language, yao2024deltazipefficientservingmultiple}, quantization~\citep{isik2023gptzip, liu2024bitdelta}, or merging~\citep{pmlr-v162-wortsman22a, yadav2023tiesmerging} to reduce the overall number of parameters. Thereby delta compression decreases both storage requirements and GPU memory utilization in scenarios involving multiple fine-tuned models.

\paragraph{Delta Parameter Decomposition} \label{sec:lowrank_approx}
Given a delta parameter $\Delta W \in \mathbb{R}^{m \times n}$, its singular value decomposition (SVD) can be expressed as $\Delta W = U \Sigma V^{\top}$, where $U \in \mathbb{R}^{m \times m}$, $V \in \mathbb{R}^{n \times n}$, and $\Sigma \in \mathbb{R}^{m \times n}$ contains the singular values in descending order. Assuming $n \leq m$ for simplicity, we can reformulate the SVD as:
\vspace{-5pt}
\begin{align}
    \Delta W &= U \Sigma V^{\top} = \sum_{i=1}^{n} \sigma_i^{\downarrow} U_i V_i^{\top}, \label{eq:svd-decomp}
\end{align}
where $U_i$ and $V_i$ denote the $i$-th columns of $U$ and $V$, respectively, and $\sigma_i^{\downarrow}$ represent the singular values ordered in descending magnitude.

We formally define the \textbf{sparsity ratio (SR)} $\alpha \in [0,1]$ as $1$ minus the ratio of the number of non-zero parameters in the sparsified delta parameters to the total number of delta parameters. 
The corresponding \textbf{compression ratio} is given by $\text{CR} = 1/(1-\alpha)$. For instance, a compression ratio of 32 corresponds to $\alpha \approx 0.97$ (97\% sparsity), yielding a 32-fold reduction in storage requirements. Through subsequent quantization, the sparse model can achieve even higher compression ratios, denoted as $\text{CR}_{\text{qt}}$ to differentiate from the sparsification-only compression ratio $\text{CR}$.

\section{Methodology}

\begin{figure*}[]
    \vspace{-10pt}
    \begin{center}
    \includegraphics[width=\textwidth]{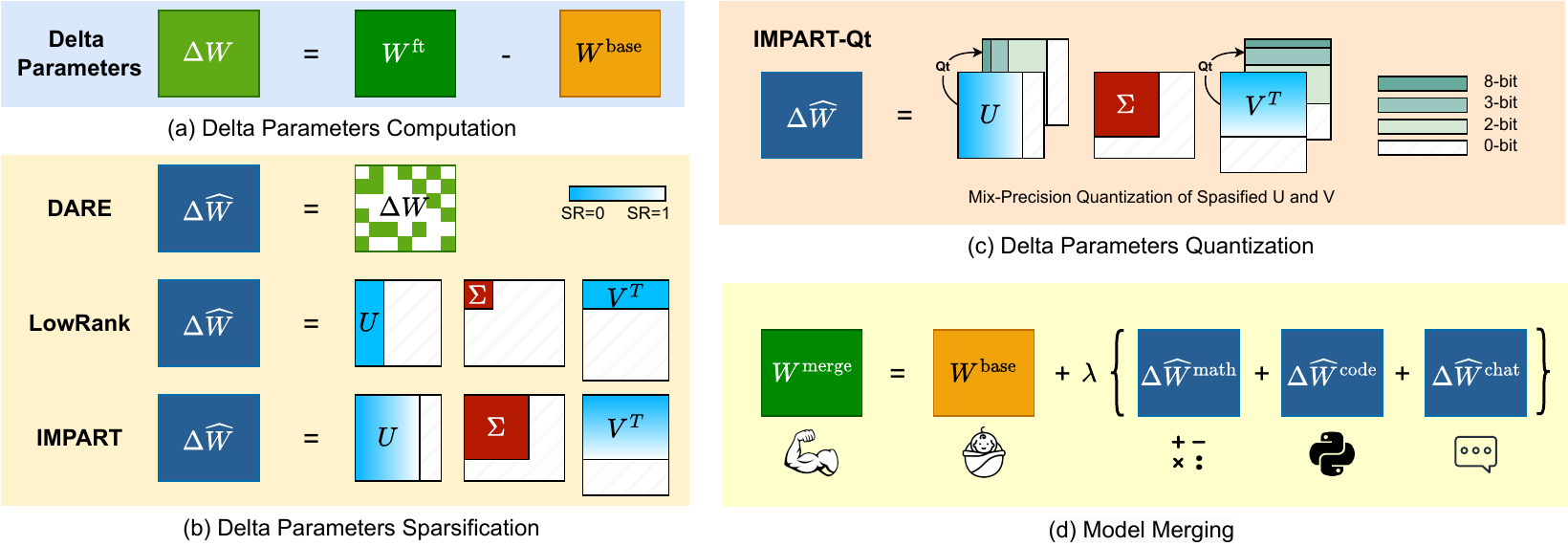}
    \vspace{-20pt}
    \caption{Overview of \mname. (a) Delta parameters computation by subtracting the base model from the fine-tuned model. (b) Comparison of delta parameters sparsification methods: DARE randomly drops delta parameters, LowRank sparsifies with low-rank approximation, and \mname adaptively sparsifies singular vectors. (c) Further apply mixed-precision quantization on sparse singular vectors to achieve higher compression ratios. (d) Model merging by combining sparsified delta parameters to build a unified multi-task model.}
    \label{fig:overview}
    \end{center}
    \vspace{-20pt}
\end{figure*}

\begin{figure}[]
    \begin{center}
    \includegraphics[width=\columnwidth]{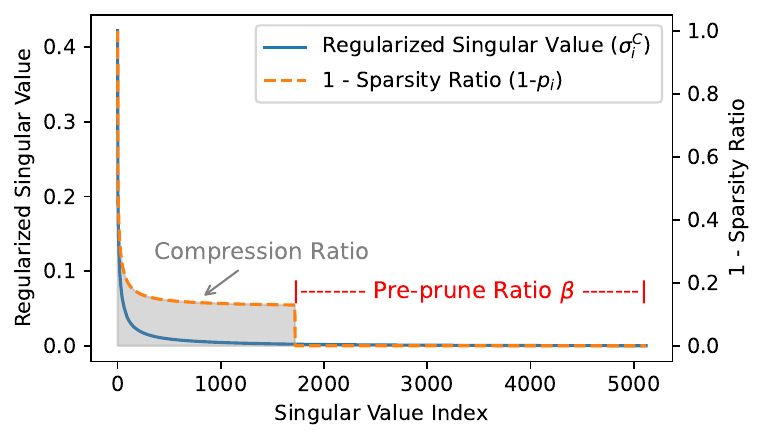}
    \vspace{-25pt}
    \caption{Importance-aware delta-sparsification adaptively sets sparsity ratios based on singular values, ensuring critical information retention. \mname first pre-prunes small singular components and then allocates sparsity budget based on regularized singular values.}
    \label{fig:mask_ratios}
    \end{center}
    \vspace{-20pt}
\end{figure}

\subsection{Importance-Aware Sparsification}

As shown in Figure~\ref{fig:overview}, \mname is an importance-aware sparsification method that adaptively allocates sparsity ratios to singular vectors based on their singular values' magnitude.
Larger singular values indicate greater importance of the corresponding singular vectors \cite{wang2025svdllm, GaoHHSJ24}, we thus assign them a smaller sparsity ratio. Conversely, singular vectors associated with smaller singular values will be given a larger sparsity ratio. Different from previous random sparsification methods like DARE \cite{yu2024language} and low-rank approximation methods \cite{ryu2023efficientstoragefinetunedmodels, saha2024compressing}, our method fully considers the importance of parameters in the SVD space, allowing for improved sparsity ratios while enhancing the performance of the sparse model.

Specifically, given $\Delta W$ with singular values $\{\sigma_k\}_{k=1}^n$, we allocate a pre-defined sparsity ratio $p_k$ (see more details in Section~\ref{sec:sr_allocation}) for the $k$-th singular vector pair ($U_k$ and $V_k$), ensuring the average sparsity ratio across all singular vectors meets the target overall sparsity ratio $\alpha$.
Inspired by the drop-and-rescale sparsification strategy in DARE \cite{yu2024language}, we then sample independent Bernoulli random variables $\xi_k^i$ and $\eta_k^j$ to randomly mask the singular vectors $U_{ik}$ (Equation~\ref{eq:u_sparse}) and $V_{kj}$ (Equation~\ref{eq:v_sparse}) according to their corresponding sparsity ratio $p_k$. To approximate the original singular vector, we apply a rescaling coefficient of $1/(1-p_k)$ to the remaining parameters (see more discussions on how to select the coefficient in Section~\ref{sec:theory}). The delta parameter is then reconstructed with sparsified singular vectors (Equation~\ref{eq:delta_sparse}).

\vspace{-10pt}
\begin{footnotesize}
\begin{align}
    \xi_k^i &\sim \text{Bernoulli}(1 - p_k), i \in [1,m], k \in [1,n]  \label{eq:xi} \\
    \eta_k^j &\sim \text{Bernoulli}(1 - p_k), k \in [1,n], j \in [1,n]  \label{eq:eta} \\
    \widehat{U}_{ik} &= U_{ik} \frac{\xi_k^i}{1-p_k}, i \in [1,m], k \in [1,n] \label{eq:u_sparse} \\
    \widehat{V}_{kj} &= V_{kj} \frac{\eta_k^j}{1-p_k}, k \in [1,n], j \in [1,n] \label{eq:v_sparse} \\
    \Delta \widehat{W} &= \widehat{U} \cdot \Sigma \cdot \widehat{V}^{\top} \label{eq:delta_sparse}
\end{align}
\end{footnotesize}

\subsection{Strategy for Sparsity Ratio Allocation} \label{sec:sr_allocation}
Given a pre-defined overall sparsity ratio $\alpha$ and $\Delta W$ with singular values $\{\sigma_k\}_{k=1}^n$, we allocate sparsity ratio $p_k$ to each singular vector pair ($U_k$ and $V_k$) based on the singular value $\sigma_k$, as shown in Figure \ref{fig:mask_ratios}:

\begin{footnotesize}
\begin{equation}
p_k =
\begin{cases}
    1 & \text{if } k > \lfloor n \cdot (1 - \beta) \rfloor \\
    \left(1 - (\frac{\sigma_k}{\sigma_1})^C\right) \cdot \gamma & \text{otherwise}
\end{cases}
\label{eq:sparse_ratio}
\end{equation}
\end{footnotesize}

\noindent where $\beta$ and $C$ are hyperparameters selected with a validation set, and $\gamma$ is a scaling factor calculated for each $\Delta W$ to ensure the overall sparsity ratio $\alpha$ is met. Our allocation strategy is designed following two key insights: 1) Previous works show that directly removing the smallest singular components can achieve performance comparable to or even better than using the full set of parameters, due to the long tail distribution of singular values \cite{ping2024deltacome, sharma2024the, ryu2023efficientstoragefinetunedmodels}. This observation motivates us to design a pre-pruning ratio, denoted as $\beta$, which aims to directly remove these long-tail singular components. 2) For the rest singular components, we allocate the sparsity ratio based on the regularized singular value, with $C$ serving as a regularization hyperparameter. In practice, it is possible that we cannot achieve the target sparsity ratio $\alpha$ by simply scaling $\gamma$ as $p_k$ is constrained to be less than $1$. We address this issue by shifting the boundary of the piecewise function to the left, thereby attaining the desired sparsity ratio. See Algorithm \ref{alg:drop_ratio} in Appendix \ref{app:sparse} for more details.\footnote{To achieve an overall sparsity ratio of $\alpha$, the sparsity ratios of $U$ and $V$ are approximately $(1+\alpha)/2$ for a square matrix.}

\subsection{Theoretical Analysis} \label{sec:theory}

We provide theoretical proof that the expectation of the reconstructed output matches the original output. Given a fine-tuned weight $W^{\text{ft}} \in \mathbb{R}^{m\times n}$ and an input $X \in \mathbb{R}^n$, the expectation of the $i$-th ($1\leq i \leq m$) dimension of the hidden state $h \in \mathbb{R}^m$ is computed as:
\vspace{-5pt}

\begin{footnotesize}
\begin{align}
    &\mathbb{E}[h_i] = \mathbb{E}[\sum_j W^{\text{ft}}_{ij}X_{j}] \notag \\
        =& \mathbb{E}[\sum_j W^{\text{base}}_{ij}X_{j}] + \mathbb{E}[\sum_j \Delta W_{ij} X_{j}]  \notag \\
        =& \sum_j W^{\text{base}}_{ij} X_{j} + \sum_j \Delta W_{ij} X_j \notag \\
        =& h_i^{\text{base}} + \sum_{j} \sum_{k} \sigma_k U_{ik}  V_{kj}  X_{j},
\end{align}
\vspace{-5pt}
\end{footnotesize}

\noindent where $h_i^{\text{base}}$ is the $i$-th dimension of the base model output. Without loss of generality, we assume that the bias term is zero.
As \mname randomly drops the $k$-th column of $U$ and $V$ independently with a sparsity ratio of $p_k$, the expectation of the reconstructed hidden state $\widehat{h}_i$ is then computed as:
\vspace{-10pt}

\begin{footnotesize}
\begin{align}
    &\mathbb{E}[\widehat{h}_i] = \mathbb{E}[\widehat{W}^{\text{ft}}X_{j}] \notag \\
        =& \mathbb{E}[\sum_j W^{\text{base}}_{ij} X_{j}] + \mathbb{E}[\sum_j \Delta \widehat{W}_{ij} X_{j}] \notag \\
        =& h_i^{\text{base}} + \mathbb{E}[\sum_{j} \sum_{k} \sigma_{k} \widehat{U}_{ik} \widehat{V}_{kj}] X_{j}       \notag \\
        =& h_i^{\text{base}} + \sum_{j} \sum_{k} \sigma_{k}  \mathbb{E}[\widehat{U}_{ik}]  \mathbb{E}[\widehat{V}_{kj}] X_{j}       \notag \\
        =& h_i^{\text{base}} + \sum_{j} \sum_{k} \sigma_{k}  [\theta \cdot (1-p_k) \cdot U_{ik}+0\cdot p_k \cdot U_{ik}] \cdot\notag \\
         & [\zeta \cdot (1-p_k) \cdot V_{kj}+0\cdot p_k \cdot V_{kj}] X_{j}       \notag \\
        =& h_i^{\text{base}} + \sum_{j} \sum_{k} \sigma_{k} [\theta \cdot (1-p_k) \cdot U_{ik}][\zeta \cdot (1-p_k) \cdot V_{kj}] X_{j}.
\end{align}
\end{footnotesize}
\vspace{-5pt}

\noindent By setting the rescaling coefficient $\theta = \zeta = 1/(1-p_k)$, we ensure that the reconstructed embedding approximates the origin. We give empirical evidence in Section~\ref{sec:ablation} to support this theoretical analysis, where the removal of the rescaling factor leads to significant performance degradation.

\section{Applications of \mname} \label{sec:method_application}
\subsection{Delta Parameter Quantization}
Previous work has demonstrated that delta parameters can be effectively compressed from 16-bits to 1-bit using low-bit quantization methods such as BitDelta~\citep{liu2024bitdelta} and Delta-CoMe~\citep{ping2024deltacome}. In this section, we enhance the compression ratio without sacrificing performance by combining \mname, a delta parameter sparsification technique, with delta parameter quantization methods. Since \mname is based on SVD, we integrate it with Delta-CoMe, a state-of-the-art mixed-precision quantization method that also operates in the SVD space. It is important to note that DARE cannot be integrated with Delta-CoMe, as SVD will break the sparse weight matrix created by DARE.

\paragraph{Delta-CoMe}
Delta-CoMe is a mixed-precision delta parameter quantization method. Instead of directly quantizing $\Delta W$, it first decomposes the delta parameter with the SVD method and then quantizes all the singular vectors using the GPTQ method~\citep{frantar2023optq}. During the process of GPTQ quantization, singular vectors corresponding to larger singular values are allocated with larger bit-widths, due to their greater impact on the approximation of delta weights.
\vspace{-5pt}
\paragraph{\sc \textbf{ImPart-Qt}}
The \mnameqt framework is a highly efficient mixed-precision delta compression method that combines the strengths of \mname and Delta-CoMe methods.
To integrate \mname with Delta-CoMe, we first use \mname to sparsify the delta parameter, and then apply Delta-CoMe to the sparsified singular vectors. However, this is not trivial. We address potential issues such as the quantization of sparse singular matrices by Delta-CoMe, the allocation of compression ratios for sparsification and quantization, and other related concerns in Appendix~\ref{app:impart_qt_implement}.

\subsection{Model Merging}\label{model_merge}
Model merging aims to merge multiple task-specific fine-tuned models into a single model with diverse abilities \cite{ilharco2023editing,yadav2023tiesmerging}. Recently, it has attracted the attention of the research community for its cost-effectiveness, knowledge-sharing potential, and space efficiency. Task Arithmetic \cite[TA]{ilharco2023editing} and TIES-Merging \cite[TIES]{yadav2023tiesmerging} are two commonly used model merging methods (see Appendix \ref{app:merge_method} for the details). As a sparsification method, \mname is able to preserve the abilities of fine-tuned LLM, as long as a small portion of the parameters in singular vectors remain unaffected. This motivates us to employ \mname before model merging, as \mname can reduce parameter redundancy in each fine-tuned model before merging,  which can potentially mitigate the interference of parameters among multiple fine-tuned models.

\begin{table*}[!t]
    \centering
    \resizebox{\textwidth}{!}{
        \begin{tabular}{lcccccccccccl}
            \toprule
            \multirow{2}{*}{\textbf{Methods}}
            & \multirow{2}{*}{CR}
            & \multicolumn{2}{c}{WizardMath-13B}
            & \multicolumn{2}{c}{WizardCoder-13B}
            & \multicolumn{2}{c}{LLaMA2-Chat-13B}
            & \multicolumn{2}{c}{LLaMA2-Chat-7B}
            & \multicolumn{2}{c}{LLaMA3-Inst-8B}
            & \multirow{2}{*}{\textbf{Avg.}} \\
            &
            & {\footnotesize GSM8K}
            & {\footnotesize MATH}
            & {\footnotesize HumanEval}
            & {\footnotesize MBPP}
            & {\footnotesize IFEval}
            & {\footnotesize AlpacaEval}
            & {\footnotesize IFEval}
            & {\footnotesize AlpacaEval}
            & {\footnotesize IFEval }
            & {\footnotesize AlpacaEval} &  \\
            \hline
            \textcolor{gray}{Backbone$^\dagger$} & \textcolor{gray}{1} & \textcolor{gray}{17.80} & \textcolor{gray}{3.90} & \textcolor{gray}{32.32} & \textcolor{gray}{62.70} & \textcolor{gray}{19.04} & \textcolor{gray}{0.71} & \textcolor{gray}{20.52} & \textcolor{gray}{0.10} & \textcolor{gray}{11.46} & \textcolor{gray}{0.08} & \textcolor{gray}{16.86} \\
                \textcolor{gray}{Fine-tuned$^\dagger$} & \textcolor{gray}{1} & \textcolor{gray}{63.96} & \textcolor{gray}{14.10} & \textcolor{gray}{59.76} & \textcolor{gray}{67.70} & \textcolor{gray}{33.64} & \textcolor{gray}{18.39} & \textcolor{gray}{31.79} & \textcolor{gray}{15.63} & \textcolor{gray}{48.80} & \textcolor{gray}{32.13} & \textcolor{gray}{38.59} \\
            \hline
            DARE & 32 & 58.91 & \textbf{11.76} & 54.27 & 64.60 & 24.77 & 2.27 & 16.82 & 0.36 & 30.50 & 17.76 & 28.20 \\
            LowRank & 32 & 56.25 & 7.94 & 57.32 & \textbf{68.80} & 26.06 & 8.45 & 23.84 & 5.72 & 29.39 & 17.18 & 30.10 \\
            \rowcolor{gray!10} \mname & 32 & \textbf{60.20} & 10.38 & \textbf{59.76} & 68.00 & \textbf{26.80} & \textbf{9.88} & \textbf{27.91} & \textbf{7.13} & \textbf{33.27} & \textbf{18.77} & \textbf{32.21} \\
            \bottomrule
        \end{tabular}
    }
\caption{Comparison of \mname and baselines on various tasks across backbones. $\dagger$ denotes the uncompressed backbone and fine-tuned models, serving as the reference for sparsification. The best results are highlighted in \textbf{bold}.}
\vspace{-15pt}
\label{tab:sparse_results}
\end{table*}

Specifically, given $N$ models fine-tuned on $N$ distinct tasks from the same base LLM, we first apply \mname on delta parameters for each fine-tuned model. Then we adopt established model merging methods such as TA and TIES to fuse the derived parameters and obtain the merged single model.
The purpose and usage of \mname are similar to the DARE method~\citep{yu2024language} in model merging, therefore, we also compare our method with DARE in Section~\ref{sec:exp_merge}.

\section{Sparsification Experiments}\label{sec:experiments}

To evaluate the effectiveness of \mname, we conduct experiments across three diverse tasks: mathematical problem-solving, code generation, and chat. Our experiments cover various model sizes and backbones, benchmarking \mname against state-of-the-art methods for model sparsification.

\subsection{Tasks}
\label{sec:tasks}
\paragraph{Mathematics} We evaluate on GSM8K~\cite{cobbe2021trainingverifierssolvemath} and MATH~\cite{hendrycks2021measuring} using Pass@1 accuracy, focusing on complex mathematical reasoning abilities.
\vspace{-6pt}
\paragraph{Code Generation} Performance is assessed on HumanEval~\cite{chen2021evaluatinglargelanguagemodels} and MBPP~\cite{austin2021programsynthesislargelanguage} using Pass@1 accuracy for natural language to code generation.
\vspace{-6pt}
\paragraph{Chat} Models are evaluated using IFEval~\cite{zhou2023instructionfollowingevaluationlargelanguage} loose prompt metric for response controllability and AlpacaEval2~\cite{dubois2024lengthcontrolledalpacaevalsimpleway} length-controlled win rate (LCWR) against GPT4-Turbo baseline, judged by GPT-4o-2024-08-06.

\subsection{Hyperparameter Selection}
For each task, we tune the hyperparameters on the validation set to select the optimal $\beta$ from \{0.6, 0.7, 0.8\} and $C$ from \{0.5, 1\}. We use SVAMP~\citep{patel-etal-2021-nlp} Pass@1, Mercury~\citep{du2024mercury} Pass@1, and FollowBench~\citep{jiang-etal-2024-followbench} hard satisfaction rate for math, code, and chat tasks as validation sets, respectively.

\subsection{Models}

The model setups are summarized in Table~\ref{tab:model_setup}.
We evaluate \mname on mainstream fine-tuned models, including WizardMath-13B-V1.0 \citep{luo2025wizardmathempoweringmathematicalreasoning} for mathematical problem solving, WizardCoder-13B \citep{luo2023wizardcoderempoweringcodelarge} for code generation, and LLaMA2-Chat-13B \citep{touvron2023llama2openfoundation} for chat tasks.
To further assess \mname’s performance across different model sizes and backbones, we also conduct experiments on LLaMA2-Chat-7B \citep{touvron2023llama2openfoundation} and LLaMA3-Instruct-8B \citep{grattafiori2024llama3herdmodels} for chat tasks.

\begin{table}[!t]
    \centering
    \resizebox{0.8\columnwidth}{!}{
        \begin{tabular}{l|l|l}
            \toprule
            \textbf{Task} & \textbf{Backbone} & \textbf{Fine-tuned} \\
            \hline
            Math & LLaMA2-13B & WizardMath-13B-V1.0 \\
            Code & Codellama-13B & WizardCoder-13B \\
            Chat & LLaMA2-13B & LLaMA2-Chat-13B \\
            Chat & LLaMA2-7B & LLaMA2-Chat-7B \\
            Chat & LLaMA3-8B & LLaMA3-Instruct-8B \\
            \bottomrule
        \end{tabular}
    }
    \caption{Selected backbones and fine-tuned LLMs for the examined tasks.}
    \vspace{-20pt}
    \label{tab:model_setup}
\end{table}

\subsection{Baselines}
\paragraph{DARE} We compare against DARE~\cite{yu2024language}, a delta compression method through random delta parameter sparsification.

\vspace{-6pt}
\paragraph{LowRank} We implement a simple SVD-based baseline~\cite{ryu2023efficientstoragefinetunedmodels} that preserves only the top $r$ singular values and corresponding singular vectors. This serves as a direct comparison point for evaluating \mname's adaptive sparsification mechanism over basic rank truncation.

\subsection{Results}

Table \ref{tab:sparse_results} presents the sparsification results for \mname and baselines across various tasks and backbones. \mname consistently outperforms both DARE and the LowRank baseline, achieving an average improvement of 4.01 over DARE and 2.11 over the LowRank baseline.

Notably, DARE exhibits significant performance degradation on chat tasks, particularly in AlpacaEval, where random sparsification leads to repetitive responses and compromises performance on LLaMA2. While the impact on IFEval is less severe due to its rule-based metrics, the overall decline underscores the limitations of random sparsification. In contrast, \mname’s adaptive strategy mitigates these issues, ensuring better retention of task-relevant knowledge and achieving more reliable results across tasks and backbones.

When comparing \mname with the LowRank baseline, we observe significant improvements in overall performance and most individual tasks.
For instance, with a compression ratio of 32, \mname only shows a 3.76 decrease on GSM8K, while LowRank exhibits a 7.71 decrease. \mname maintains performance on HumanEval without any degradation, while LowRank exhibits a 2.44 decrease. These results underscore the effectiveness of \mname in preserving critical task-specific information and achieving SOTA model sparsification.

\section{Analyses of \mname}
\begin{table}[]
    \centering
    \resizebox{1\columnwidth}{!}{
    \begin{tabular}{cl|cccc}
        \toprule
        \textbf{ID} & \textbf{Ablations} & GSM8K & HumanEval & IFEval & \textbf{Avg.} \\ \hline
        \rowcolor{gray!10}
        \textcircled{1} & \mname & \textbf{60.20} & \textbf{59.76}& \textbf{26.80} & \textbf{48.92}\\ 
        \textcircled{2} & w/o Pre-prune & 57.92 & 54.88& 26.62 & 46.47\\
        \textcircled{3} & w/o Importance-Aware & 0.00 & 51.83& 12.20 & 21.34 \\
        \textcircled{4} & w/o Pre-prune, w/o I.A. & 0.00& 20.12& 11.83& 10.65\\
        \textcircled{5} & w/o 1/(1-p) Rescale & 33.21& 6.10 & 22.92& 20.74\\ \bottomrule
    \end{tabular}
}
\vspace{-10pt}
\caption{Ablation study on different components of \mname. I.A. denotes Importance-Aware, and $1/(1-p)$ rescale refers to the rescale coefficient in Equation~\ref{eq:u_sparse}, ~\ref{eq:v_sparse}.}
\label{tab:ablations}
\vspace{-10pt}
\end{table}

In this section, we conduct comprehensive analyses of \mname on three representative tasks: mathematical problem solving (GSM8K with WizardMath-13B), code generation (HumanEval with WizardCoder-13B), and chat (IFEval with LLaMA2-Chat-13B). Unless otherwise specified, we use a compression ratio $\text{CR}=32$.

\subsection{Ablations}\label{sec:ablation}

To assess the impact of different components of \mname, we conduct an ablation study on the pre-pruning parameter $\beta$, the importance-aware sparsification strategy, and the effectiveness of the $1/(1-p)$ rescale, as shown in Table~\ref{tab:ablations}.

Our results show that all design components contribute to \mname. First, pre-pruning long-tail singular vectors with pre-pruning ratio $\beta$ results in a more effective sparsity allocation strategy, enhancing the performance of the final sparse model by an average of 2.45 (ID 2 vs. ID 1).

We next evaluate our importance-aware sparsification strategy. In \mname, we adaptively assign sparsity ratios to singular vectors based on their importance values. Comparing this approach against uniform sparsification across unpruned singular vectors (ID 3), we observe that disregarding importance leads to a substantial performance degradation of 27.58 on average (ID 3 vs. ID 1). This deterioration worsens when pre-pruning is removed, with performance dropping by 38.27 (ID 4 vs. ID 1). Most notably, uniform sparsification produces severely degraded outputs with repetition and incoherence, resulting in complete failure (0.00 accuracy) on GSM8K. These findings demonstrate that importance-aware sparsification is crucial for preserving model capabilities.

Finally, we verify the effectiveness of the $1/(1-p)$ rescale in approximating the original model. When the rescale coefficient is removed from Equation~\ref{eq:u_sparse} and ~\ref{eq:v_sparse}, we observe a significant performance decrease of 28.18 on average (ID 5 vs. ID 1).

\subsection{Sensitivity Analysis on $\beta$ and $C$}\label{sec:hyperparam}

\begin{table}[]
    \centering
    \resizebox{\columnwidth}{!}{
        \begin{tabular}{l|c|cc|cccc}
        \toprule
        \textbf{Methods} & CR & $C$ & $\beta$ & GSM8K & HumanEval & IFEval & \textbf{Avg.} \\ \hline
        \textcolor{gray}{Backbone$^\dagger$} & \textcolor{gray}{1} & \multicolumn{2}{c|}{-} & \textcolor{gray}{17.80} & \textcolor{gray}{32.32} & \textcolor{gray}{19.04} & \textcolor{gray}{23.05}\\
        \textcolor{gray}{Fine-tuned$^\dagger$} & \textcolor{gray}{1} & \multicolumn{2}{c|}{-} & \textcolor{gray}{63.96} & \textcolor{gray}{59.76} & \textcolor{gray}{33.64} & \textcolor{gray}{63.81} \\   \hline
        DARE & 32 & \multicolumn{2}{c|}{-} & 58.91 & 54.27 & 24.77 & 45.98 \\
        LowRank & 32 & \multicolumn{2}{c|}{-} & 56.25 & 57.32 & 26.06 & 46.54 \\ \hline
        \multirow{6}{*}{\mname} & \multirow{6}{*}{32} & \multicolumn{1}{c|}{\multirow{3}{*}{0.5}} & 0.6 & 56.48 & 56.71 & \textbf{27.91} & 47.03 \\
         &  & \multicolumn{1}{c|}{} & 0.7 & 58.07 & 54.88 & 25.88 & 46.28 \\
         &  & \multicolumn{1}{c|}{} & 0.8 & 57.62 & 54.27 & \underline{26.80} & 46.23 \\ \cline{3-8}
         &  & \multicolumn{1}{c|}{\multirow{3}{*}{1}} & 0.6 & \textbf{\underline{60.20}} & \textbf{\underline{59.76}} & 26.43 &  \textbf{48.80} \\
         &  & \multicolumn{1}{c|}{} & 0.7 & 58.45 & 56.71 & 27.54 & 47.57 \\
         &  & \multicolumn{1}{c|}{} & 0.8 & 58.45 & 59.15 & 25.14 & 47.58 \\ \bottomrule
        \end{tabular}
    }
    \vspace{-10pt}
    \caption{Hyperparameter study across different tasks. The best performance is shown in \textbf{bold}, and results selected by the validation set are \underline{underlined}.}
    \vspace{-10pt}
    \label{tab:Hyperparam_diff_model}
\end{table}

We conduct a comprehensive sensitivity analysis to evaluate the impact of the pre-pruning parameter $\beta$ and regularization parameter $C$. Table \ref{tab:Hyperparam_diff_model} presents results across diverse tasks, demonstrating \mname's robust performance across various hyperparameter configurations.

Our analysis of the regularization parameter $C$ reveals task-specific effects. For mathematical reasoning and code generation tasks, maintaining the original singular values ($C=1$) yields better performance. In contrast, the chat model benefits from a smaller $C$ ($C=0.5$), which reduces the differences between regularized singular values.

\begin{table}[!]
    \centering
    \resizebox{\columnwidth}{!}{
        \begin{tabular}{l|l|ccccc}
            \toprule
            \textbf{Tasks} & Method & 8 & 16 & 32 & 64 & \textbf{Avg.} \\ \hline
            \multirow{3}{*}{GSM8K}
            & DARE & \textbf{61.79} & 60.20 & 56.63 & 53.68 & 58.08 \\
            & LowRank & 61.41 & 58.38 & 56.25 & 50.42 & 56.62 \\
            & \cellcolor{gray!10}{\mname} & \cellcolor{gray!10}{61.64} & \cellcolor{gray!10}{\textbf{62.40}} & \cellcolor{gray!10}{\textbf{60.20}} & \cellcolor{gray!10}{\textbf{56.56}} & \cellcolor{gray!10}{\textbf{60.20}} \\ \hline

            \multirow{3}{*}{HumanEval}
            & DARE & 58.54 & 58.54 & 56.71 & 57.32 & 57.78 \\
            & LowRank & 54.27 & 55.49 & 57.32 & 56.71 & 55.95 \\
            & \cellcolor{gray!10}{\mname} & \cellcolor{gray!10}{\textbf{59.15}} & \cellcolor{gray!10}{\textbf{60.37}} & \cellcolor{gray!10}{\textbf{59.76}} & \cellcolor{gray!10}{\textbf{57.93}} & \cellcolor{gray!10}{\textbf{59.30}} \\ \hline

            \multirow{3}{*}{IFEval}
            & DARE & 28.84 & 26.99 & 19.04 & 8.87 & 20.94 \\
            & LowRank & 25.32 & 27.36 & 26.06 & 24.95 & 25.92 \\
            & \cellcolor{gray!10}{\mname} & \cellcolor{gray!10}{\textbf{29.02}} & \cellcolor{gray!10}{\textbf{27.91}} & \cellcolor{gray!10}{\textbf{26.80}} & \cellcolor{gray!10}{\textbf{26.25}} & \cellcolor{gray!10}{\textbf{27.50}} \\ \bottomrule
        \end{tabular}

    }
    \vspace{-10pt}
    \caption{Performance of \mname with different compression ratios.}
    \vspace{-15pt}
    \label{tab:diff_sprase_ratio}
\end{table}

\begin{table*}[thb]
    \centering
    \resizebox{\textwidth}{!}{
        \begin{tabular}{lcccccccccccl}
            \toprule
            \multirow{2}{*}{\textbf{Methods}}
            & \multirow{2}{*}{$\text{CR}_{\text{qt}}$}
            & \multicolumn{2}{c}{{WizardMath-13B}}
            & \multicolumn{2}{c}{{WizardCoder-13B}}
            & \multicolumn{2}{c}{{LLaMA2-Chat-13B}}
            & \multicolumn{2}{c}{{LLaMA2-Chat-7B}}
            & \multicolumn{2}{c}{{LLaMA3-Inst-8B}}
            & \multirow{2}{*}{\textbf{Avg.}} \\
            &
            & {\footnotesize GSM8K}
            & {\footnotesize MATH}
            & {\footnotesize HumanEval}
            & {\footnotesize MBPP}
            & {\footnotesize IFEval}
            & {\footnotesize AlpacaEval}
            & {\footnotesize IFEval}
            & {\footnotesize AlpacaEval}
            & {\footnotesize IFEval }
            & {\footnotesize AlpacaEval} &  \\
            \hline
                \textcolor{gray}{Backbone$^\dagger$} & \textcolor{gray}{1} & \textcolor{gray}{17.80} & \textcolor{gray}{3.90} & \textcolor{gray}{32.32} & \textcolor{gray}{62.70} & \textcolor{gray}{19.04} & \textcolor{gray}{0.71} & \textcolor{gray}{20.52} & \textcolor{gray}{0.10} & \textcolor{gray}{11.46} & \textcolor{gray}{0.08} & \textcolor{gray}{16.86} \\
                \textcolor{gray}{Fine-tuned$^\dagger$} & \textcolor{gray}{1} & \textcolor{gray}{63.96} & \textcolor{gray}{14.10} & \textcolor{gray}{59.76} & \textcolor{gray}{67.70} & \textcolor{gray}{33.64} & \textcolor{gray}{18.39} & \textcolor{gray}{31.79} & \textcolor{gray}{15.63} & \textcolor{gray}{48.80} & \textcolor{gray}{32.13} & \textcolor{gray}{38.59} \\
            \hline
            BitDelta & 32 & 61.11 & 12.12 & 51.83 & 58.50 & 25.32 & \textbf{18.30} & 27.36 & 11.87 & 34.38 & 26.35 & 32.71 \\
            \dareqt & 32 & 62.17 & 13.40 & 57.32 & 67.70 & \textbf{30.87} & 17.68 & \textbf{29.76} & 11.76 & 42.33 & 28.24 & 36.12 \\
            Delta-CoMe & 32 & 62.40 & 12.56 & 56.71 & 68.30 & 27.91 & 15.52 & 29.39 & 10.85 & 41.40 & 26.64 & 35.17 \\
             \rowcolor{gray!10} \mnameqt & 32 & \textbf{64.29} & \textbf{13.54} & \textbf{58.54} & \textbf{68.50} & \textbf{30.87} & 17.65 & \textbf{29.76} & \textbf{12.55} & \textbf{45.84} & \textbf{28.27} & \textbf{36.98} \\
            \bottomrule
        \end{tabular}
    }
    \vspace{-8pt}
\caption{Comparison of \mnameqt and baselines on various tasks across backbones. $\dagger$ denotes the uncompressed backbone and fine-tuned models, serving as the reference for quantization.
$\text{CR}_{\text{qt}}$ denotes the combined compression ratio with sparsification and quantization. The best results are highlighted in \textbf{bold}.}
\vspace{-12pt}
\label{tab:quant_results}
\end{table*}

Regarding the pre-pruning ratio $\beta$, we find that moderate values ($\beta=0.6$) typically yield optimal results, striking a balance between removing noise and retaining important information. Higher values ($\beta=0.7, 0.8$) lead to marginally decreased performance, suggesting the loss of important task-specific knowledge during aggressive pre-pruning.

When analyzing the validation set’s selections, we find that it effectively identifies near-optimal hyperparameters for math and code-related tasks but exhibits limitations for chat tasks. For instance, it selects a configuration that achieves 26.80 on IFEval, falling short of the optimal 27.91, likely due to misalignment between validation and test set. Despite this suboptimal configuration, \mname still outperforms all baselines on chat tasks, highlighting its robustness and effectiveness in model sparsification.

\subsection{Different Compression Ratios}\label{sec:diff_sparse_ratio}
To demonstrate the flexibility of \mname, we evaluate performance across varying compression ratios ($8$ to $64$). Table~\ref{tab:diff_sprase_ratio} (visualized in Figure~\ref{fig:diff_sparse_ratio}) demonstrates that \mname consistently outperforms baseline methods across most settings, with its advantages becoming more pronounced at higher compression ratios. These results validate \mname's effectiveness in preserving task-specific knowledge under aggressive sparsification.

\section{Applications of \mname}

\subsection{Delta Parameter Quantization}\label{sec:exp_quant}

\paragraph{Setup}
We compare \mnameqt with three baselines: BitDelta, DARE-Qt, and Delta-CoMe, by evaluating them with the same model and benchmark setup as in Section \ref{sec:experiments}.
We set the target compression ratio $\text{CR}_{\text{qt}}$ to 32 for all tasks and models.
In line with Delta-CoMe, we employ a triple-precision quantization scheme, assigning 8-bit, 3-bit, and 2-bit precision to distinct singular value groups. See Appendix~\ref{app:sparsify_quant_allocation} for more details. 
\vspace{-5pt}
\paragraph{Results}
Table \ref{tab:quant_results} presents the quantization results for different quantization methods. \mnameqt achieves the highest overall performance, with an average score of 36.98, surpassing BitDelta by 4.27, \dareqt by 0.86, and Delta-CoMe by 1.81. These results highlight the effectiveness of \mnameqtc’s adaptive sparsification strategy in preserving essential task-specific parameters while achieving a high compression ratio. Compared to uncompressed aligned models, \mname achieves near-lossless performance on math and code tasks. However, there is a relatively greater performance degradation on chat tasks. This suggests that the difficulty of compression varies across different types of tasks. Compared to the sparsification results in Table~\ref{tab:sparse_results}, \mnameqt achieves significantly better outcomes than \mname at the same compression ratio of 32. This indicates that for effective compression of the delta parameter, a combination of sparsification and quantization is preferable to using either method alone.

We further present the performance across varying compression ratios, ranging from $16$ to $128$. As shown in Table~\ref{tab:diff_quant_ratio} (visualized in Figure~\ref{fig:diff_quant_ratio} of Appendix~\ref{app:qtp} ), \mnameqt consistently outperforms baseline methods across most compression settings.

\begin{table}[]
    \centering
    \resizebox{\columnwidth}{!}{
        \begin{tabular}{l|l|ccccc}
            \hline
            \textbf{Tasks} & Method & 16 & 32 & 64 & 128 & \textbf{Avg.} \\ \hline
            \multirow{4}{*}{GSM8K}
            & BitDelta & 59.89 & 61.11 & 61.11 & 59.14 & 60.31 \\
            & \dareqt & 62.55 & 62.17 & 62.09 & 58.00 & 61.20 \\
            & Delta-CoMe & 61.94 & 62.40 & 61.62 & 58.23 & 61.05 \\
            & \cellcolor{gray!10}{\mnameqt} & \cellcolor{gray!10}{\textbf{64.22}} & \cellcolor{gray!10}{\textbf{64.29}} & \cellcolor{gray!10}{\textbf{62.32}} & \cellcolor{gray!10}{\textbf{60.35}} & \cellcolor{gray!10}{\textbf{62.80}} \\ \hline

            \multirow{4}{*}{HumanEval}
            & BitDelta & 52.44 & 51.83 & 51.22 & 50.00 & 51.37 \\
            & \dareqt & 61.59 & 57.32 & 56.71 & 55.49 & 57.78 \\
            & Delta-CoMe & 59.15 & 56.71 & 52.44 & 55.49 & 55.95 \\
            & \cellcolor{gray!10}{\mnameqt} & \cellcolor{gray!10}{\textbf{62.20}} & \cellcolor{gray!10}{\textbf{58.54}} & \cellcolor{gray!10}{\textbf{57.32}} & \cellcolor{gray!10}{\textbf{56.71}} & \cellcolor{gray!10}{\textbf{58.69}} \\ \hline

            \multirow{4}{*}{IFEval}
            & BitDelta & 25.88 & 25.32 & 23.66 & 22.92 & 24.45 \\
            & \dareqt & 31.79 & \textbf{30.87} & 28.65 & \textbf{27.73} & 29.76 \\
            & Delta-CoMe & 31.24 & 27.91 & 28.10 & 25.88 & 28.28 \\
            & \cellcolor{gray!10}{\mnameqt} & \cellcolor{gray!10}{\textbf{32.16}} & \cellcolor{gray!10}{\textbf{30.87}} & \cellcolor{gray!10}{\textbf{30.68}} & \cellcolor{gray!10}{\textbf{27.73}} & \cellcolor{gray!10}{\textbf{30.36}} \\ \hline
        \end{tabular}
    }
    \vspace{-5pt}
    \caption{Performance of \mnameqt with different compression ratios $\text{CR}_{\text{qt}}$.}
    \vspace{-10pt}
    \label{tab:diff_quant_ratio}
\end{table}

\subsection{Model Merging}
\label{sec:exp_merge}

\paragraph{Setup}

We evaluate model merging on three representative benchmarks for math, code, and chat tasks, including GSM8K, HumanEval, and IFEval. We use WizardMath-13B and LLaMA2-Chat-13B as the mathematical and chat-specialized models that are fine-tuned from LLaMA2.
Since model merging requires fine-tuned models sharing the same backbone, we fine-tune the LLaMA2-13B backbone on the Magicoder dataset ~\cite{wei2024magicoderempoweringcodegeneration} to obtain the code specialized model, which we refer to as LlamaCoder. The detailed fine-tuning configuration is shown in the Appendix \ref{app:our_code_model}.
We integrate \mname into two common merging strategies: TA and TIES, and compare \mname with DARE and no pre-sparsification. Please refer to Appendix \ref{app:merge_method} for more details.

\vspace{-5pt}
\paragraph{Results}

Table \ref{tab:merging_results} summarizes the merging results for \mname across various tasks and merging strategies. \mname achieves the highest average scores of 40.98 and 39.99 for TA and TIES, outperforming DARE by 0.46 and 0.78.
Compared to no re-sparsification, \mname improves model merging performance by 0.47 and 1.71 for TA and TIES, respectively. In contrast, DARE shows minimal improvement in TA merging performance. These results underscore the effectiveness of \mname in improving model merging.

\begin{table}[]
    \centering
    \resizebox{\columnwidth}{!}{
        \begin{tabular}{l|c|c|ccc|c}
            \toprule
            \textbf{Models} & Merge & Mask & GSM8K & HumanEval & IFEval & \textbf{Avg.} \\ \hline
            Math & - & No & 63.96 & - & - & - \\
            Code & - & No & - & 52.44 & - & - \\
            Chat & - & No & - & - & 33.64 & - \\ \hline
            \multirow{6}{*}{\makecell{Chat\&\\Math\&\\Code}} & \multirow{3}{*}{TA} & No & 62.02 & 30.49 & 29.02 & 40.51 \\
             &  & DARE & 61.26 & \textbf{31.10} & \textbf{29.21} & 40.52 \\
             &  & \cellcolor{gray!10}{\mname} & \cellcolor{gray!10}{\textbf{63.00}} & \cellcolor{gray!10}{\textbf{31.10}} & \cellcolor{gray!10}{28.84} & \cellcolor{gray!10}{\textbf{40.98}} \\ \cline{2-7}
             
             & \multirow{3}{*}{TIES} & No & 57.54 & 24.39 & 32.90 & 38.28 \\
             &  & DARE & \textbf{59.59} & 24.39 & 33.64 & 39.21 \\
             &  & \cellcolor{gray!10}{\mname} & \cellcolor{gray!10}{58.45} & \cellcolor{gray!10}{\textbf{26.22}} & \cellcolor{gray!10}{\textbf{35.30}} & \cellcolor{gray!10}{\textbf{39.99}} \\
             \bottomrule
        \end{tabular}
    }
    \vspace{-8pt}
    \caption{Comparison of different sparsification strategies for model merging.}
    \vspace{-16pt}
    \label{tab:merging_results}
\end{table}

\section{Related Work}\label{sec:related_work}

\paragraph{Model Sparsification}
The increasing size of LLMs has made model compression a critical research focus. While traditional model pruning approaches \citep{ijcai2018p330, lee2021layeradaptive} remove parameters based on magnitude, they often lead to significant performance degradation when applied to fine-tuned models \citep{yao2024deltazipefficientservingmultiple}. Recent work has instead focused on delta-sparsification, where ERE \citep{ryu2023efficientstoragefinetunedmodels} employs low-rank decomposition of delta weights, and DARE \citep{yu2024language} demonstrates the effectiveness of random parameter dropping. However, these methods either disregard parameter importance entirely or evaluate it at too coarse a granularity. In contrast, \mname introduces importance-aware sparsification that assesses and prunes individual singular vectors, achieving superior performance.

\paragraph{Model Quantization}
Parameter quantization has emerged as a prominent compression technique, with GPTQ \citep{frantar2023optq} pioneering error-minimizing low-bit-width approaches. Subsequent innovations have extended to mixed-precision quantization across model weights \citep{dettmers2023spqrsparsequantizedrepresentationnearlossless}, activations \citep{shen2023agilequantactivationguidedquantizationfaster}, and layers \citep{bablani2024efficienteffectivemethodsmixed}. In the context of delta parameters, initial approaches like GPT-Zip \cite{isik2023gptzip} and DeltaZip \citep{yao2024deltazipefficientservingmultiple} achieved 2-bit compression through GPTQ extensions and structured pruning, while BitDelta \citep{liu2024bitdelta} advanced to 1-bit compression using trainable scaling factors. Delta-CoMe \cite{ping2024deltacome} further enhanced efficiency by introducing varying bit-width representations for singular vectors. \mname builds upon these advances by integrating importance-aware sparsification with Delta-CoMe, establishing new SOTA compression performance.

\paragraph{Model Merging}
The proliferation of task-specific models \citep{luo2025wizardmathempoweringmathematicalreasoning, luo2023wizardcoderempoweringcodelarge, wei2024magicoderempoweringcodegeneration} from open-source pre-trained backbones \citep{touvron2023llama2openfoundation, grattafiori2024llama3herdmodels, jiang2023mistral7b} has motivated efficient model merging techniques to reduce deployment costs. While initial approaches like parameter averaging \citep{pmlr-v162-wortsman22a, ilharco2023editing} demonstrated the potential of combining delta parameters, subsequent methods addressed parameter conflicts through Fisher information matrices \citep{matena2022mergingmodelsfisherweightedaveraging}, linear regression \citep{jin2023dataless}, and magnitude-based parameter selection \citep{yadav2023tiesmerging}. Although DARE \citep{yu2024language} introduced random delta weight dropping during merging, it overlooks parameter importance. \mname advances this direction by incorporating importance-aware sparsification in the SVD space, leading to more effective model merging.

\section{Conclusion}

We introduced \mname, a novel importance-aware delta-sparsification approach for efficient model compression and merging in large language models. By leveraging singular value decomposition to adaptively determine sparsity ratios based on parameter importance, \mname effectively preserves critical task-specific knowledge while achieving significant sparsification. Our comprehensive experiments in mathematical reasoning, code generation, and chat tasks demonstrate that \mname consistently outperforms existing sparsification methods. Additionally, \mname can be integrated with state-of-the-art delta-quantization and model merging techniques, achieving new benchmarks in both delta-quantization and model merging.

\section*{Limitations}

While we demonstrate the effectiveness of \mname in compressing and merging LLMs, several limitations remain. First, \mname treats all weight matrices equally and does not consider the potential benefits of layer-wise pruning, which have been shown to improve compression performance and model efficiency \citep{lee2021layeradaptive, li2024setar, dumitru2024changeconstantdynamicllm, wang2025adaptpruneradaptivestructuralpruning, li2025adasvdadaptivesingularvalue}. Future work could explore fine-grained sparsification strategies for different layers and weight matrices to further enhance compression performance. Second, \mname requires a validation set to determine the optimal hyperparameters.
Despite this being a common practice in model compression \citep{frantar2023optq, ping2024deltacome}, it may not always lead to the optimal model due to the potential misalignment between the validation and test sets. Nevertheless, \mname consistently achieves state-of-the-art performance across multiple tasks and various hyperparameter configurations, demonstrating its robustness.

\bibliography{custom}

\clearpage
\appendix

\section{More Details for \mname}
\label{app:sparse}
\subsection{Sparsity Ratio Allocation}
Algorithm \ref{alg:drop_ratio} shows the details of sparsity ratio allocation across singular vectors for a given target ratio of $\alpha$. For simplicity, we only present the case of square matrices.

\begin{algorithm}[H]
    \footnotesize
    \caption{Sparsity Ratios Computation}
    \label{alg:drop_ratio}
    \begin{algorithmic}[1]
        \Require{Singular values $\{\sigma_i\}_{i=1}^n$, Target sparsity ratio $\alpha$, Pre-prune ratio $\beta$, Rescale parameter $C$}
        \Ensure{Sparsity ratio list $P$ for singular vectors in $U$ and $V$}
        \State $\alpha \gets (1+\alpha)/2$  \Comment{Update sparsity ratio for $U$ and $V$}
        \State Let $r = \lfloor n \cdot (1 - \beta) \rfloor$
        \For{$i \gets r+1$ \textbf{to} $n$}
            $p_i \gets 1$ \Comment{Pre-prune}
        \EndFor
        \State  $\gamma \gets \text{min}(\frac{\alpha-\beta}{1-\beta} \cdot \frac{r}{\sum_{i=1}^r(1-(\frac{\sigma_i}{\sigma_1})^C)}, \frac{1}{(1-(\frac{\sigma_r}{\sigma_1})^C} )$
        \For{$i \gets 1$ \textbf{to} $r$}
            $p_i \gets (1 - (\frac{\sigma_i}{\sigma_1})^C) \cdot \gamma$ \Comment{Importance-aware sparsification}
        \EndFor
        \State $i \gets r$
        \While{$\frac{1}{r} \sum_{k=1}^{r} p_k < \alpha$}
            \State $p_i \gets 1$ \Comment{Shift boundary to meet target sparsity ratio}
            \State $i \gets i - 1$
        \EndWhile
        \State \Return $P \gets \{p_k\}_{k=1}^n$
    \end{algorithmic}
\end{algorithm}

\section{More Details for \mnameqt}
\label{app:impart_qt_implement}
\subsection{GPTQ for Sparse Weight}
\label{app:sparse_gptq}
Delta-CoMe quantizes the left and right singular matrix using GPTQ with a designed mix-precision strategy. However, GPTQ has been primarily confined to dense models. We extend GPTQ to accommodate sparse matrices. Specifically, during the column-by-column quantization process, we apply a sparsification mask to the parameters, ensuring that only those retained after sparsification are subject to quantization. Furthermore, when updating the remaining weights based on quantization error, we compute the error solely on the retained parameters. The detailed algorithm is presented in Algorithm \ref{alg:sparse_gptq}. 

\begin{algorithm}[hbt]
\footnotesize
\caption{GPTQ for Sparse Weight}
\label{alg:sparse_gptq}
\begin{algorithmic}[1]
\Require{Weight to be quantized $W$ and its corresponding mask $M$, Inverse Hessian $H^{-1}=(2XX^T+\lambda I)^{-1}$, and blocksize $B$}
\Ensure{Quantized weight $Q$}
\State Initialize $\mathbf{Q} \gets \mathbf{0}_{d_\text{row} \times d_\text{col}}$ \Comment{Quantized output}
\State Initialize $\mathbf{E} \gets \mathbf{0}_{d_\text{row} \times B}$ \Comment{Block quantization errors}
\State $\mathbf{H}^{-1} \gets \text{Cholesky}(\mathbf{H}^{-1})$ \Comment{Hessian inverse information}
\For{$i=0, B, 2B,\dots$}
    \For{$j = i, \dots, i + B - 1$}
        \State $\mathbf{M}_{\text{tmp}} \gets \mathbf{M}_{:,j}$
        \State $\mathbf{W}_{\text{tmp}} \gets \mathbf{W}_{:,j} \odot \mathbf{M}_{\text{tmp}}$
        \Comment{Set the sparsified weight to zero}
        \State $\mathbf{Q}_{:,j} \gets \text{quant}(\mathbf{W}_{\text{tmp}}) \odot \mathbf{M}_{\text{tmp}}$
        \State $\mathbf{E}_{:, j - i} \gets (\mathbf{W}_{\text{tmp}} - \mathbf{Q}_{:, j}) \, / \, [\mathbf{H}^{-1}]_{jj}$
        \Comment{Quantization error}
        \State $\mathbf{W}_{:, j:(i + B)} \gets \mathbf{E}_{:, j - i} \cdot \mathbf{H}^{-1}_{j, j:(i + B)}$
        \Comment{Update weights in block}
    \EndFor
    \State $\mathbf{W}_{:, (i + B):} \gets \mathbf{E} \cdot \mathbf{H}^{-1}_{i:(i + B), (i + B):}$
    \Comment{Update all remaining weights}
\EndFor
\end{algorithmic}
\end{algorithm}

\subsection{Compression Ratio Allocation}\label{app:sparsify_quant_allocation}
In line with Delta-CoMe, we employ a triple-precision quantization scheme, assigning 8-bit, 3-bit, and 2-bit precision to distinct singular value groups. The first group consists of 2 elements, the second group includes 32 elements, and the remaining elements form the third group. To achieve the target compression ratio $\text{CR}_{\text{qt}}$ after quantization, the corresponding sparsity ratio $\alpha$ is calculated using a binary search process, as described in Algorithm~\ref{alg:find_drop_ratio}. For simplicity, we only present the case of square matrices.

\begin{algorithm}[H]
    \footnotesize
    \caption{Binary Search to Find Overall Sparsify Ratio for Compression with Quantization}
    \label{alg:find_drop_ratio}
    \begin{algorithmic}[1]
    \Require{Singular values $\{\sigma_i\}_{i=1}^n$, Compression ratio $\text{CR}_{\text{qt}}$, Pre-prune ratio $\beta$, Rescale parameter $C$, Tolerance $\text{tol}$}
    \Ensure{Overall sparsify ratio $\alpha$}
    \State $\text{low} \gets 0$, $\text{high} \gets 1$
    \Comment{Set the lower and upper bound}

    \While{$\text{high} - \text{low} > \text{tol}$}
        \State $\text{mid} \gets 0.5 \cdot (\text{low} + \text{high})$ \Comment{Compute the midpoint}

        \State $P \gets \text{Algorithm~\ref{alg:drop_ratio}}(\{\sigma_i\}_{i=1}^n, \text{mid}, \beta, C)$

        \Comment{Compute the sparsity ratios $P$ using Algorithm~\ref{alg:drop_ratio}}

        \State $\alpha_{\text{qt}} \gets \frac{1}{n}(\frac{1}{2} \sum_{i=1}^{2} (1-p_i) + \frac{1}{4} \sum_{i=3}^{34} (1-p_i) + \frac{1}{8} \sum_{i=35}^{r} (1-p_i))$

        \Comment{Calculate sparsification ratio after quantization}

        \If{$\frac{1}{1-\alpha_{\text{qt}}} < 2\cdot\text{CR}_{\text{qt}}$}
            \State $\text{low} \gets \text{mid}$ \Comment{Update lower bound}
        \Else
            \State $\text{high} \gets \text{mid}$ \Comment{Update upper bound}
        \EndIf
    \EndWhile
    \State \Return $0.5 \cdot (\text{low} + \text{high})$
    \end{algorithmic}
\end{algorithm}

\subsection{The Storage of Sparsification Mask}

Technically, a random sparsification of $U$ (Equation~\ref{eq:u_sparse}) and $V$ (Equation~\ref{eq:v_sparse}) would necessitate storing sparsity masks for reconstruction (Equation~\ref{eq:delta_sparse}), resulting in additional storage overhead. To address this issue, we implement a deterministic seeding strategy: we initialize the random seed for $\xi_k^i$ (Equation~\ref{eq:xi}) using $\sigma_k$ when sparsifying $U$, and use random seed+1 for $\eta_k^j$ (Equation~\ref{eq:eta}). This approach maintains the independence of $\xi_k^i$ and $\eta_k^j$ while enabling the reconstruction of sparsity masks directly from the singular value $\sigma_k$, thus avoiding additional storage.

\begin{figure*}[t]
    \centering
    \begin{subfigure}[b]{0.32\textwidth}
        \centering
        \includegraphics[width=\textwidth]{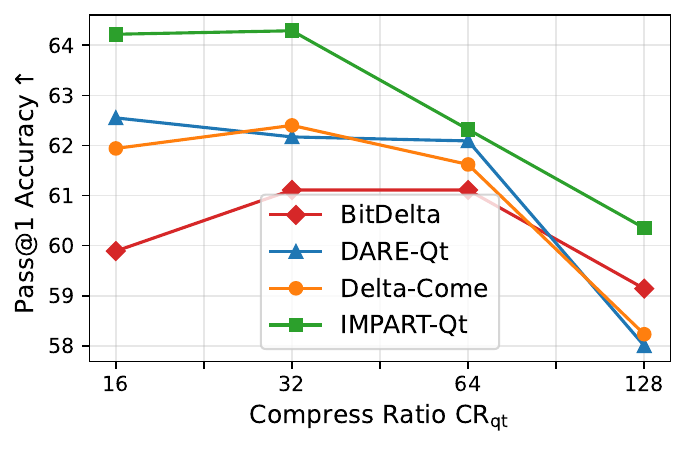}
        \caption{WizardMath-13B on GSM8K}
        \label{fig:diff_quant_ratio_gsm8k}
    \end{subfigure}
    \hfill
    \begin{subfigure}[b]{0.32\textwidth}
        \centering
        \includegraphics[width=\textwidth]{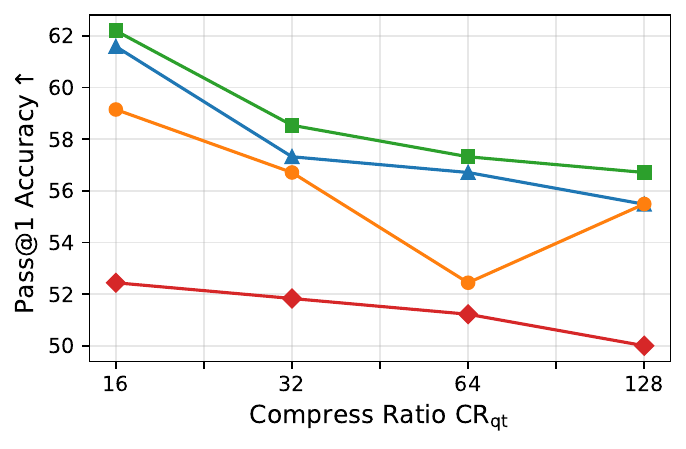}
        \caption{WizardCoder-13B on HumanEval}
        \label{fig:diff_quant_ratio_humaneval}
    \end{subfigure}
    \hfill
    \begin{subfigure}[b]{0.32\textwidth}
        \centering
        \includegraphics[width=\textwidth]{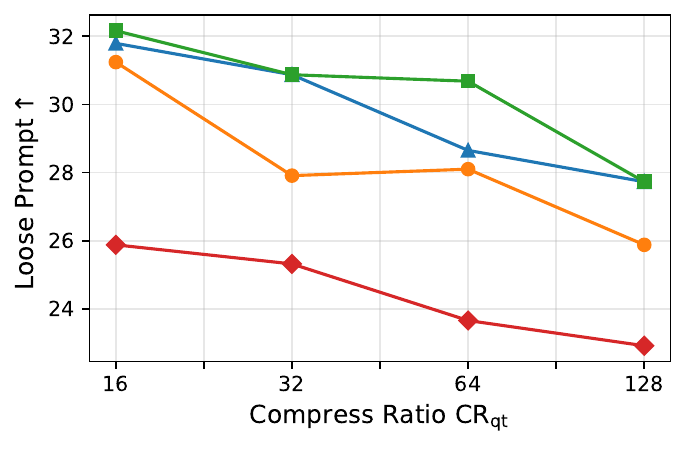}
        \caption{LLaMA2-Chat-13B on IFEval}
        \label{fig:diff_quant_ratio_ifeval}
    \end{subfigure}
    \caption{Comparative evaluation of \mname against state-of-the-art quantization methods across mathematical reasoning, code generation, and chat tasks (more detailed discussions are in Section~\ref{sec:exp_quant}).}
    \label{fig:diff_quant_ratio}
\end{figure*}

\subsection{Baselines for \mnameqt}

\paragraph{BitDelta} BitDelta~\cite{liu2024bitdelta} achieves a 1/16 compression ratio by compressing the task vector into $\mu\odot\mathrm{Sign}(\Delta)$, where $\mathrm{Sign}(\cdot)$ denotes the 1-bit element-wise sign of each parameter and $\mu$ is a trainable scaling factor. In this paper, we further combine BitDelta with DARE to achieve an even higher compression ratio.

\paragraph{\dareqt}
\dareqt is the baseline that integrates DARE into GPTQ. DARE first sparsifies the delta parameters, and then GPTQ further quantizes the sparsified delta parameters. To quantize the sparse delta parameters, we use the same version of GPTQ as shown in Appendix~\ref{app:sparse_gptq}.
For each compression ratio $\text{CR}_{\text{qt}}$, we use GPTQ to quantize the 16-bit parameters into  2/4/8-bit, with the sparsity ratio $\alpha$ of DARE determined by the target compression ratio $\text{CR}_{\text{qt}}$. Then we report the configuration that achieved the best performance on the validation set for each compression ratio $\text{CR}_{\text{qt}}$.

\paragraph{Delta-CoMe} We faithfully implement Delta-CoMe as described in the original paper~\cite{ping2024deltacome}, achieving the target compression ratio by adjusting the number of 2-bit singular vectors.

\subsection{Performance of \mnameqt Across Compression Ratios}
\label{app:qtp}
Figure \ref{fig:diff_quant_ratio} visualizes the performance of \mnameqt and baselines on different tasks across compression ratios of $16$ to $128$.

\section{More Details for Model Merging}
\label{app:merge_method}
\subsection{Common Model Merging Methods}
\paragraph{TA}
Task Arithmetic~\cite{ilharco2023editing} leverages a scaling term to regulate the contributions of the pre-trained backbone and the aggregated delta parameters set, formed by summing $n$ multiple individual delta parameters:

\vspace{-20pt}
\begin{small}
\begin{align}
    W^{\text{merge}} = W^{\text{base}} + \lambda * \sum_{t=1}^n \Delta W^{\text{t}},
\end{align}
\end{small}
\vspace{-20pt}

\paragraph{TIES}
TIES-Merging~\cite{yadav2023tiesmerging} aims to address parameter conflicts in model merging. Given a delta parameters set, it first trims parameters with lower magnitudes,

\vspace{-15pt}
\begin{small}
\begin{align}
    \Delta W^{\text{t}} = trim(\Delta W^{\text{t}}).
\end{align}
\end{small}
\vspace{-15pt}

Then, TIES elects the sign with the highest total magnitude to resolve sign disagreements:

\vspace{-15pt}
\begin{small}
\begin{align}
    \gamma^t = \textrm{sgn}(\Delta W^{\text{t}}),\\
    \gamma^m = \textrm{sgn}(\sum_{t=1}^{n} \Delta W^{\text{t}}).
\end{align}
\end{small}
\vspace{-15pt}

Finally, Parameters with consistent signs are disjointly merged:

\vspace{-15pt}
\begin{small}
\begin{align}
    \mathcal{A} = {\{t \in [n] ~|~ \gamma^t = \gamma^{merge}\}},\\
    W^{\text{merge}} = W^{\text{base}} + \lambda * \frac{1}{|\mathcal{A}|}\sum_{t \in \mathcal{A}} \Delta W^{\text{t}}.
\end{align}
\end{small}
\vspace{-15pt}

\subsection{Details of LlamaCoder}
\label{app:our_code_model}
We implement LlamaCoder by full fine-tuning from the Llama2-13B base model using the Magicoder dataset~\cite{wei2024magicoderempoweringcodegeneration}. The training process involved 3 epochs, with a batch size of 8, a peak learning rate of 2e-5, and a maximum sequence length of 4096.
Note that we do not use WizardCoder-13B as its backbone is Codellama-13B.

\subsection{Hyperparameter Selection}
We follow DARE \cite{yu2024language,NEURIPS2024_99fc8bc4} for hyperparameter search. Specifically, we perform a grid search to optimize the hyperparameters of TA and TIES. Specifically, for both methods, the scaling term is selected from the set \{0.4, 0.6, 0.8, 1.0, 1.2\}, and for TIES, the retain ratio of the largest-magnitude parameters is chosen from \{0.4, 0.6, 0.8\}.
When incorporating the sparsification methods DARE and \mname into TA/TIES, we use the pre-selected hyperparameters of TA/TIES and search for the optimal sparsification ratios from \{0.1, 0.3, 0.5, 0.7, 0.9\} to save computation.

\end{document}